# *Underwater Autonomous Tank Cleaning Rover*

Aditya Sundarajan
School of Electronics Engineering
*Vellore Institute of Technology*
Chennai, India.
aditya.sundarajan2019@vitstudent.ac.in

Kevin Timothy Muller
School of Electronics Engineering
*Vellore Institute of Technology*
Chennai, India.
kevintimothy.muller2019@vitstudent.ac.in

Jaideepnath Anand S
School of Electronics Engineering
*Vellore Institute of Technology*
Chennai, India.
jaideepnath.anands2019@vitstudent.ac.in

Mangal Das
School of Electronics Engineering
*Vellore Institute of Technology*
Chennai, India.
mangal.das@vit.ac.in

***Abstract* - In order to keep aquatic ecosystems safe and healthy, it is imperative that cleaning be done frequently. This research suggests the use of autonomous underwater rovers for effective underwater cleaning as a novel approach to this issue. The enhanced sensing and navigational capabilities of the autonomous rovers enable them to independently navigate underwater environments and find and remove underwater garbage and uneaten fish feed which can be recycled. The suggested solution not only does away with the requirement for human divers, but also provides a more effective and affordable technique for underwater cleaning. The paper also examines the creation, testing, and potential of the autonomous underwater rovers.**

***Keywords* -** *Autonomous , ROS , SLAM*

## I. INTRODUCTION

The wellbeing of both people and marine life depends on the health and safety of aquatic ecosystems. Underwater cleaning, which entails removing trash and pollutants from the water, is one of the most crucial activities for sustaining these environments. Traditional underwater cleaning techniques, however, can be risky for human divers and are frequently time- and money-consuming. The creation of autonomous underwater cleaning devices is gaining popularity as a solution to these problems [6] , [7]. These systems detect and remove debris on their own, without the assistance of a human, using cutting-edge sensors and control systems.
This research explores the conception and use of an autonomous underwater cleaning system with a view to its effectiveness, efficiency, and potential environmental impact. The paper will specifically look into the propulsion, sensing, control, and cleaning technologies that make up an autonomous underwater cleaning system. It will also look at the opportunities and difficulties of deploying autonomous systems for underwater cleaning, including their ability to navigate challenging conditions and work continuously.

The papers' ultimate goals are to add to the increasing body of knowledge about autonomous underwater systems and to offer insightful information about how autonomous underwater cleaning could be used to enhance the health and safety of aquatic environments. A crucial duty that can help reduce these risks is underwater cleaning, but conventional methods [1] can be time-consuming , expensive, and frequently require a high level of risk for human divers. The development of autonomous underwater cleaning devices, which can find and remove debris automatically without requiring human interaction, has attracted increasing interest as a solution to these problems.

This paper also explores the development and application of an autonomous underwater cleaning system, with a particular emphasis on the use of artificial intelligence (AI) [2] , [4] , [8] and machine learning (ML) approaches to improve the system's performance. The paper will investigate the use of sophisticated sensors and control systems to allow the system to identify and categorize various forms of trash, as well as the creation of clever algorithms for streamlining cleaning procedures and reducing energy usage.

In addition to its technological contributions, the paper will look at how autonomous underwater cleaning systems might affect society, the economy, and the environment.
Using autonomous systems for underwater cleaning offers the potential to increase cleaning operations' efficiency and safety, save costs, and lessen the environmental impact of human participation.

## II. SOFTWARE

One of the software used in this project is ROS - Robot Operating System. For a fully autonomous robot with path planning and obstacle-avoiding capabilities, we use some important components of ROS, namely SLAM and navigation node, among others.

### *A. SLAM NODE*

The SLAM node is a package that produces a map using a virtual LIDAR sensor present in the node. Using this node enables users to create maps based on the virtual environment the robot is being placed in and enables efficient path planning in later stages of development.
In simple terms, the SLAM node produces a 2D map of the desired environment and saves it natively on the machine, which can be called by the navigation node for path planning and autonomous navigation set by the user.

### *B. ROS GAZEBO*

This package in ROS enables the construction of custom virtual environments to test robot mobility. It also contains many default environments for basic testing.In this simulation software, users are able to test their robot movements and add near-real-life environment simulations and make changes to their initial design, thereby reducing hardware costs and runtime errors which can occur with hardware testing.

Additionally, Gazebo will publish to ROS /clock topic, an in-built timekeeping node that tracks simulation time, to give user simulation data.

For our project, we use gazebo to create a virtual environment, as shown in Fig. 2 similar to a pool or the actual operating environment the rover will be placed into, and connect this world to RVIZ and provide a map for user-defined path planning.

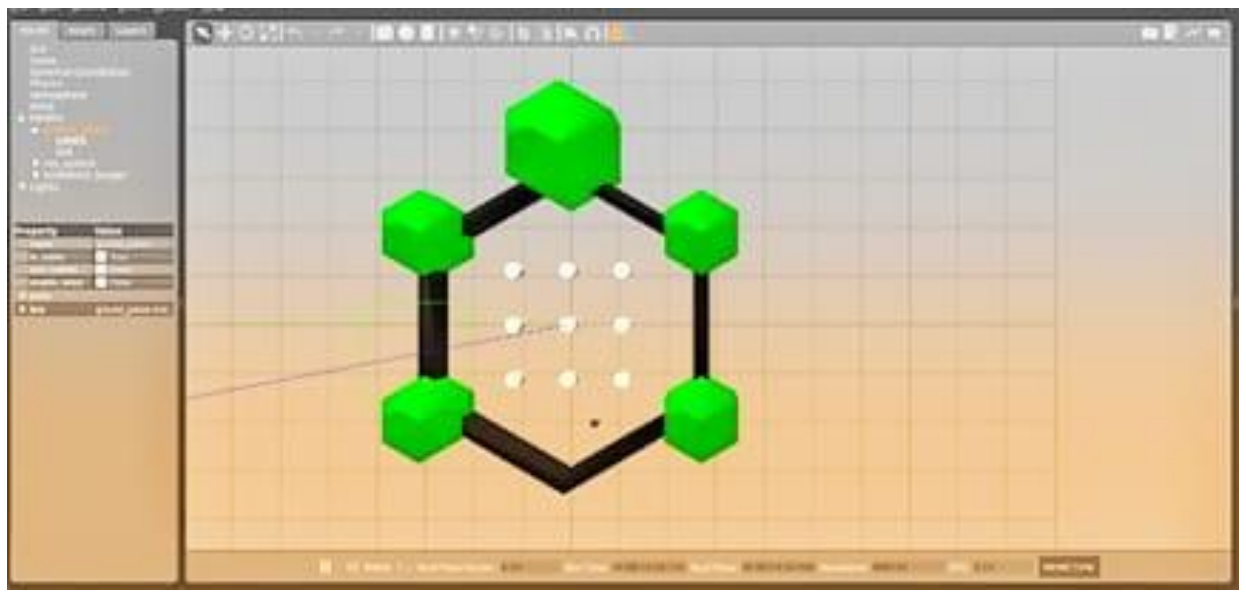

Fig. 1. ROS Gazebo turtleworld

### *C. RVIZ*

RVIZ is a powerful visualization tool generally used for robots and algorithms. It augments ROS Visualization. Using this tool the user/developer can envision how a robot perceives its world in both real and simulated environments. It can also update you on the current location and state of the robot. This is done by getting inputs such as sensor data and enabling it to visualize a more precise picture of the machine's perception of its locality.

We use RVIZ to map the virtual environment created in gazebo and save the virtual map, as shown in Fig 2. which will be later used in the web application for the user to mark the path along which the rover will move through.

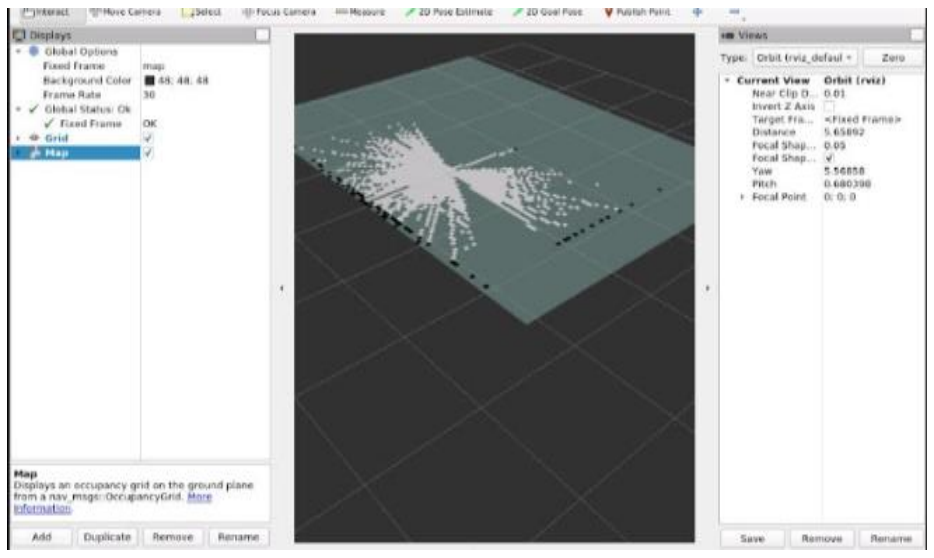

Fig. 2. ROS Rviz Mapping virtual environment

### *D. VISUAL STUDIO CODE*

VS is a popular code editor known for its flexibility and compatibility with various coding languages. This software is used to mainly design the front end of our application in which the user has the privilege of marking the path to be cleaned by the rover.

A web application [5] which interacts with ROS workspace to get a user defined path for the robot is built using VS Code. The LIDAR scanned map from the virtual environment in ROS workspace is imported here and the user defines the path along which the robot is programmed to move through. In Fig. 3, the coordinates chosen by the user are marked in red by the web application.

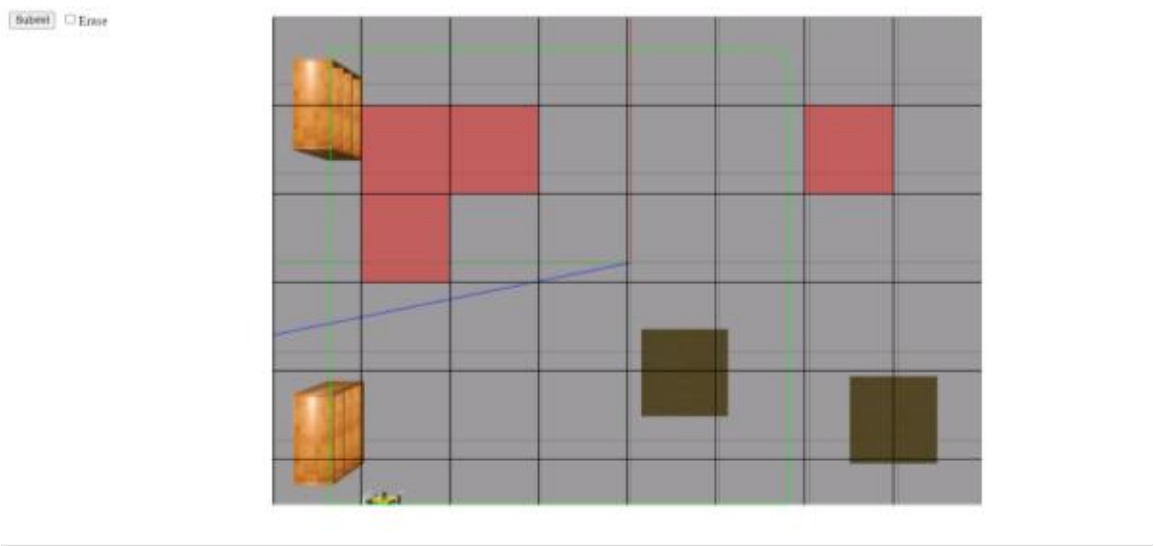

Fig. 3. Web application for user to define path

The coordinates are programmed to show on a return link with other details such as total size of pool and size of rover as shown in Fig 4.

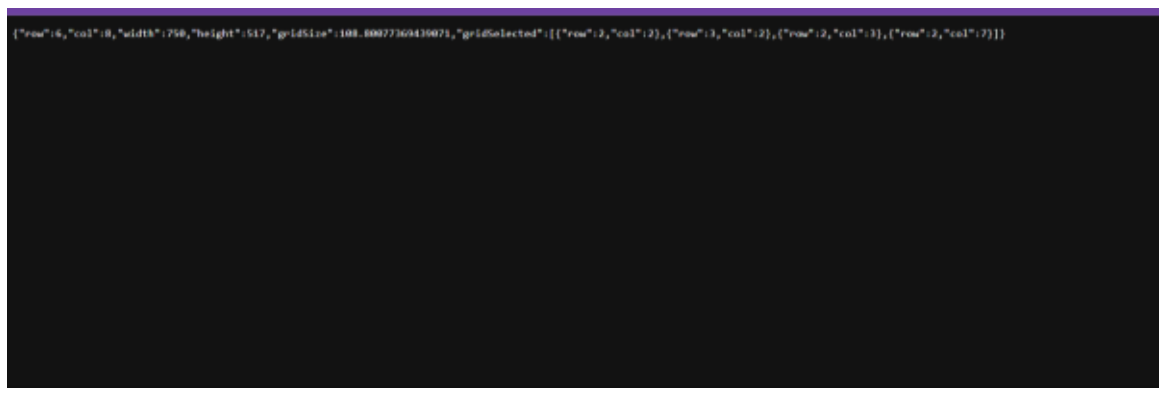

Fig. 4. JSON instance output given by website

The successful implementation of this web application enables one to selectively clean parts of their pool or aquatank which otherwise would not be possible.

## III. WORKING PRINCIPLE

For the mapping and user-defined path, we have created a two-part software model.
We create a virtual world similar to the environment the rover will be operating in and run SLAM [3] node in ROS which provides us with a workable map of the environment.

To get user input in path planning, we export the map produced by SLAM node into a web application we have created using Visual Studio, which displays the map to the user, where a path can be plotted and exported to the ROS workspace which communicates said commands to the controller.

## IV. DESIGN APPROACH

The rover model scans a virtual environment custom-built to mimic the environment the robot will be placed in real-life. Once the respective nodes are run in the ros terminal, we can map the virtual world and create a navigable map for the web application to accept as input and the user can define a path for the rover to follow.

The mapping stage requires the user to manually control the rover to get an accurate map. This process is done only once and the remainder of the rover operation is autonomous. The user gets JSON output as coordinates which are inserted back into ROS for the robot to follow.

## V. OBSERVATION

Once our simulations are running, we are able to import the coordinates set by the user for the web application into the VS compiler and send them as input coordinates to the ros navigation stack for the global path planner package to plot a path into rviz as per the given coordinates.

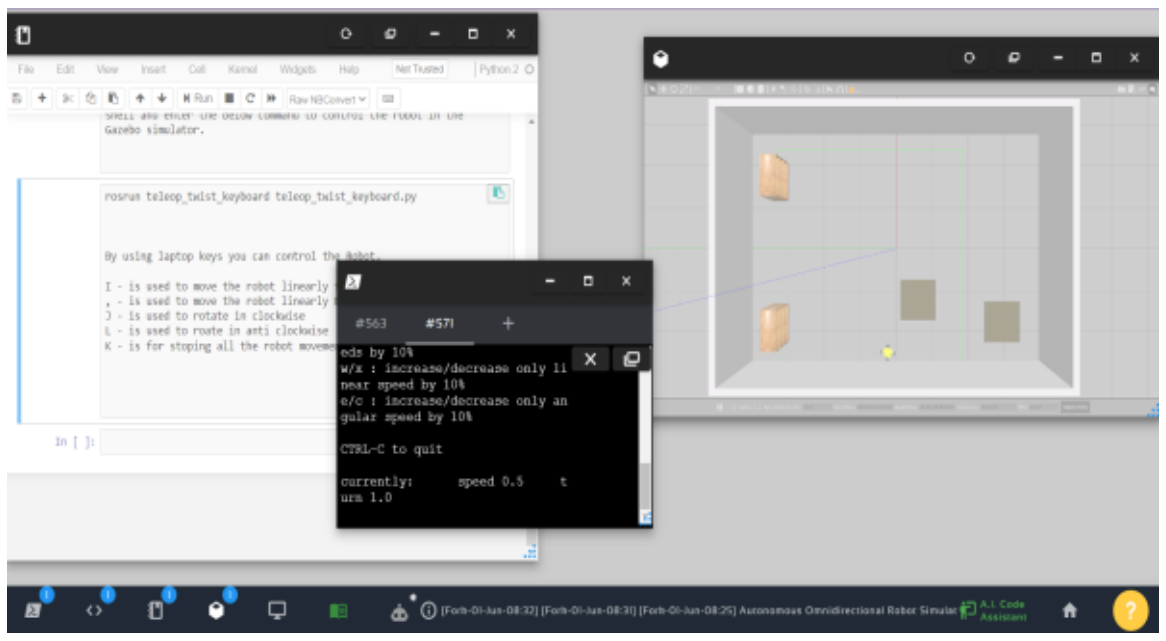

Fig. 5. Software Simulation

We are able to import the mapped file to the website. The website then projects the image and allows the user to set the path for the robot to follow. Once the path is set, the user can finish their input and the coordinates are generated. These coordinates are then imported into the path planning algorithm present in the navigation node in ROS, which accepts input from the website and plots the path onto RVIZ and we are able to visualize the robot moving in the programmed path in Gazebo virtual world as seen in Fig. 5.

With localization prebuilt and run in SLAM node, the robot moves based on the input coordinates, in the path set by the user.

## VI. RESULTS AND INFERENCE

During the progression of this project, the software phase had shown consistent success with respect to the algorithm. This algorithm was structured in ROS as stated above which shows the rover on a self - learnt map which when fed ,the end 2D-navigation goal , finds or rather learns an efficient reachable 'path' or 'way' to the end destined goal avoiding all obstacles and immobile areas within the map.

Ultimately , the entire purpose of the problem statement for finding an efficient (unclean) path to clean has been satisfied by the rover by making use of the coordinates given by the Web Application / User.

## VII. CONCLUSION

This paper presents the implementation of a software model of autonomous navigation of an underwater rover which is capable of receiving predetermined path planning from users and autonomously navigating the environment without any user intervention. This has been implemented using ROS Gazebo, where the virtual environment is produced, and Rviz, which is responsible for creating 2D maps for users to plan the path for the robot to follow. The navigation packages used to achieve this outcome are SLAM and global path planning respectively. Additionally, an external path planning web application was implemented to receive the scanned map form ros workspace and produce user defined coordinates which are used as input for ros path planning package to plot the path in rviz.

## ACKNOWLEDGEMENT

We would like to extend our endless gratitude to all our supporters and contributors who have helped us complete this paper. Firstly, we would like to thank Mangal Das ,our guide for this project, for his valuable insights into the subject and his support and encouragement which led to the completion of this paper. We would also like to thank Karthik S from Anna University Coimbatore, who was instrumental towards the completion of the project. Finally, we thank VIT University for providing us with the facilities to complete the paper. Without the support from these individuals, this paper would not have been possible.